%% file: main.tex
\documentclass[10pt,letterpaper,conference]{article}

\usepackage[utf8]{inputenc}

\usepackage{sectsty}
\sectionfont{\fontsize{14}{16}\selectfont}
\subsectionfont{\itshape}
\subsubsectionfont{\itshape}

\usepackage{multicol}
\usepackage[a4paper, margin=2cm]{geometry}
\usepackage[pdftex]{graphicx}
\usepackage{float}
\usepackage[hang,small,bf,tight]{subfigure}
\usepackage[font=small]{caption}
\usepackage{amsmath}
\usepackage{amssymb}
\usepackage{array,multirow}

\usepackage{indentfirst}
\usepackage{hyperref}
\hypersetup{colorlinks,allcolors=black}
\usepackage{dirtytalk}
\begin{document}
\twocolumn[
  \begin{@twocolumnfalse}
    \title{Tension Estimation and Localization \\ for a Tethered Micro Aerial Robot}
    \author{Ricardo Martins and Meysam Basiri}
    \date{}
    \maketitle
\end{@twocolumnfalse}
]

%Corpo
\input{1abstract.tex}
\input{Draft.tex}
\input{DraftM.tex}

\input{DraftE.tex}
\input{8conclusion.tex}

%Biblio e style
\bibliographystyle{unsrt}
\bibliography{Bib}

\end{document}

%% file: 1abstract.tex
\begin{abstract}
\textbf{
     This work focuses on the study of tethered fights of a micro quadcopter, with the aim of supplying continuous power to a small-sized aerial robot. Multiple features for facilitating the interaction between a tethered micro quadcopter and a ground base are described in this paper. Firstly, a tether model based on the catenary curve is presented that describes a quadcopter tethered to a point in space. Furthermore, a method capable of estimating the tension applied to the quadcopter, based only on the inertial information from the IMU sensors and the motor thrusts, is presented. Finally, a novel method for localizing the quadcopter by exploiting the tension imposed by the tether and the shape of the tether is described. The proposed methods are evaluated both in simulation and in real world prototype.
     }
\end{abstract}

%% file: Draft.tex
\section{Introduction}{

Micro aerial robots are known by their versatility and ability to get information from higher altitudes allowing them to have many important applications \cite{basiriJFR, pimentel2022bimodal,zefri_inspection_2017,kayan_heat_2018}.  However, such robots have a low payload capacity and due to the small on-board battery short flight times are reached. In some applications where the robot is only required to operate inside a limited air space, such as to inspect a solar panel installation \cite{zefri_inspection_2017}, an industrial structure \cite{burri_aerial_2012} or a vessel \cite{ortiz_vessel_2014} or to operate inside a house \cite{10.1007/978-3-319-70833-1_64}, continuous power can be supplied trough a tether that is attached to a ground station allowing beyond battery missions \cite{bib:iros}. The ground station can also be mobile, such as a mobile service robot \cite{ventura2016domestic} or an unmanned ground vehicle \cite{basiri2021multipurpose}, to further extend the operation area of the tethered aerial robot and to perform missions cooperatively with the aerial/ground multi robot system \cite{9811264}. The integration of a tethered micro aerial robot can also enhance the capabilities of mobile ground robots by providing an aerial perspective of the surrounding environment, facilitating tasks such as path planning and obstacle avoidance \cite{9827047,9561062}. 

To implement a tethered solution, the cable characteristics for the tether must be carefully considered taking into account the resistance and the varying power drop across the cable \cite{bib:v_drop,bib:platform_tethered}. Although it is possible to only power an aerial robot through a tether, however, a small on-board battery could also be used to allow robustness against power blackouts or damages to the tether \textbf{\cite{bib:switch_APU}}.  

%Given that ground-robots have a large payload capacity that allows them to carry large batteries, sensors and computers, they can be employed alongside aerial robots in many different applications to perform tasks in a collaborative manner \cite{basiri2021multipurpose}. In exchange, aerial robots can also provide the ground robots with important aerial sensing and help improve their localization. In such collaborative scenarios, ground robots could act as mobile charging stations \cite{s23020829} or mobile power units that power the aerial robot through an attached tether \cite{bib:construction, bib:awareness}. 
Ground robots, with their ability to carry large payloads, including batteries, sensors, and computers, offer a unique opportunity for collaboration with aerial robots \cite{basiri2021multipurpose}. By combining their strengths, the two types of robots can undertake tasks that would otherwise be impossible individually. Aerial robots, for example, can provide valuable aerial sensing and enhance the ground robots' localization capabilities \cite{s23020829}. Conversely, ground robots can serve as mobile charging stations or power sources, supplying energy to their aerial counterparts via an attached tether \cite{s23020829,bib:construction, bib:awareness}

%The quadcopter can also be attached to a non-movable ground-base, which does not have the same versatility as if it was attached a ground-robot, but still allows the completion of complex tasks in a more efficient way \cite{bib:iros}. Moreover, the tethered flights can be extended to cooperative missions between multiple quadcopters, where they are attached between them or to a heavy load to solve transportation problems \cite{bib:tognon} \cite{bib:multicopters}.
\par
Despite the benefits mentioned above, there are multiple challenges that must be considered to facilitate autonomous operation of tethered aerial robots.  The constraints in the motion of the robot imposed by the tether and the varying tension that the tether applies to the robot must be considered by the flight controller. For this purpose, it is important to have an accurate estimation of this varying tension. 

In this paper a method to estimate the tension on a tethered aerial robot is described that is only based on the IMU sensor readings. As the IMU measurements have high noise levels which prevents them from being used directly for accurate tension estimations, a filtering approach is used to assist with the estimation.  Furthermore, we show that the robot position estimation can be improved by exploiting the properties of the tether and using the tension applied to the UAV. The tension on the end-points of the tether is related to its shape, which means that the relation between the tether model and the tension applied to the quadcopter can be expressed mathematically.
}

%% file: DraftM.tex
\section{Proposed Methodologies}
\subsection{Characterization of the shape of the tether}
\label{sec:t_shape}
\subsubsection{Catenary Curve}
\begin{figure}[h]
    \centering
    \includegraphics[width=0.7\linewidth]{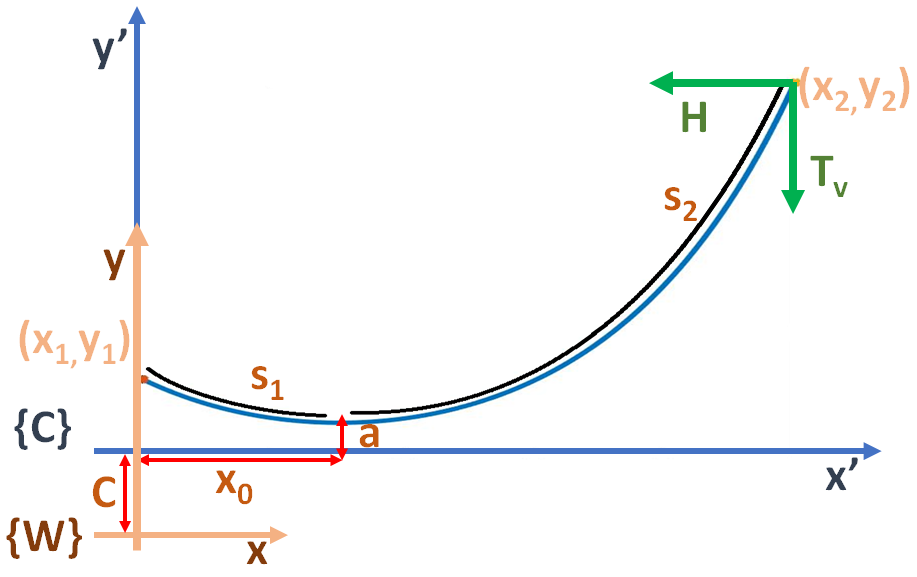}
    \caption{Catenary curve and related parameters; world $\{W\}$ and catenary $\{C\}$ frames.}
    \label{fig:cat_curve}
\end{figure}

In case of a non-rigid tether connecting the quadcopter to a ground-base, the shape of the tether approximately outlines a catenary curve. The catenary model consists of a hanging cable, with no stiffness, sagging under its own weight and supported only by its ends (see figure \ref{fig:cat_curve}).
The point ($x_1$, $y_1$) corresponds to the origin and the point ($x_2$, $y_2$) to the position of the quadcopter. The shape of the catenary can be defined according to a mathematical model, in which expression \ref{eq:cat_expression} presents the \textbf{equation of the catenary}.
\begin{equation}
\label{eq:cat_expression}
    y=a.cosh\Big(\frac{x-x_0}{a}\Big)+C
\end{equation}
Parameter $\langle x_0\rangle$ is the abscissa of the lowest point. Parameter $\langle a\rangle$ corresponds to the $y$ coordinate of the lowest point of the curve ($x=x_0$) regarding the catenary frame $\{C\}$, and it must always be positive.\footnote{A negative value of parameter $\langle a\rangle$ would only have a physical meaning if the shape of the catenary was concave instead of convex.}. Parameter $\langle C\rangle$ is an offset between the world frame $\{W\}$ and the catenary frame $\{C\}$, which depends on the tether's parameter $\langle a\rangle$ and the $y$ coordinate of the lowest point regarding the world frame ($y_0$).
\begin{equation}
    \label{eq:Cparam}
    C=y_0-a
\end{equation}
\par
\begin{figure}[h]
    \centering
    \includegraphics[width=0.6\linewidth]{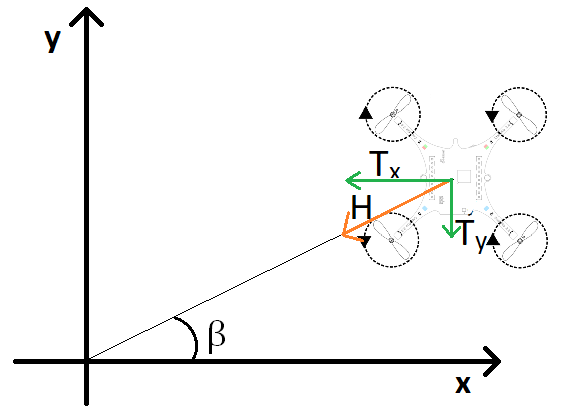}
    \caption{Horizontal tension decomposition.}
    \label{fig:t0_txty}
\end{figure}

Expressions \ref{eq:cat_s1} and \ref{eq:cat_s2} introduce the tether parameters $\langle s_1\rangle$ and $\langle s_2\rangle$, which represent the arc-length from the tether lowest point to the origin and to the UAV, respectively.
\begin{equation}
\label{eq:cat_s1}
    s_1=a.sinh\Big(\frac{|x_1-x_0|}{a}\Big)
\end{equation}
\begin{equation}
\label{eq:cat_s2}
    s_2=a.sinh\Big(\frac{|x_2-x_0|}{a}\Big)
\end{equation}
Equations \ref{eq:cat_T0} and \ref{eq:cat_Tv} respectively present the horizontal and vertical tension on the end-points of the catenary curve. Both the horizontal and vertical tension depend on the tether's parameters - $\langle a\rangle$ or $\langle s\rangle$ - and on the weight of the tether, where $\omega$ is the weight per length unit \cite{bib:cat_book}.
\begin{equation}
\label{eq:cat_T0}
    H=\omega.a
\end{equation}
\begin{equation}
\label{eq:cat_Tv}
    T_V=\omega.s
\end{equation}
The absolute value of the tension results from the euclidean norm of the horizontal and vertical tensions.
\begin{equation}
\label{eq:cat_absTension}
    |T|=\sqrt{T_V^2+H^2}
\end{equation}
The quadcopter flying in an $R^3$ space allows to define the horizontal tension ($H$) in terms of a component along the $x$ direction and another along the $y$ direction, as shown in figure \ref{fig:t0_txty}.
The previous circumstance relates the tension along the \textit{x} and \textit{y} direction according equation \ref{eq:Tx} and \ref{eq:Ty}, respectively.
\begin{equation}
\label{eq:Tx}
    Tx=cos(\beta).|H|
\end{equation}
\begin{equation}
\label{eq:Ty}
    Ty=sin(\beta).|H|
\end{equation}
\par
The parameters of the catenary cannot be mathematically computed by only knowing the coordinates of the two tether end-points. Thus, these parameters - $\langle a\rangle$, $\langle x_0\rangle$, $\langle C\rangle$, $\langle s_1\rangle$, and $\langle s_2\rangle$ - are obtained gathering additional information from knowing the full length of the tether.
\par
This assumption leads to a possible and determined system of equations \ref{eq:sys2_eq1}-\ref{eq:sys2_eq5}, which can be numerical solved generating the curve parameters. 
\begin{equation}
\label{eq:sys2_eq1}
    y_1=a.cosh(\frac{x_1-x_0}{a})+C
\end{equation}
\begin{equation}
\label{eq:sys2_eq2}
    y_2=a.cosh(\frac{x_2-x_0}{a})+C
\end{equation}
\begin{equation}
\label{eq:sys2_eq3}
    s_{total}=s_2+s_1
\end{equation}
\begin{equation}
\label{eq:sys2_eq4}
    s_1=a.sinh\Big(\frac{|x_1-x_0|}{a}\Big)
\end{equation}
\begin{equation}
\label{eq:sys2_eq5}
    s_2=a.sinh\Big(\frac{|x_2-x_0|}{a}\Big)
\end{equation}
Subtracting equation \ref{eq:sys2_eq1} from equation \ref{eq:sys2_eq2}, and making use of the hyperbolic cosine properties, it follows:
\begin{equation}
\label{eq:cat4}
    \Delta Y= 2a.sinh(\frac{\Delta x}{a}).sinh(\frac{x_{average}-x_0}{a}),
\end{equation}
where $\Delta x=\frac{x2-x1}{2}$, $x_{average}=\frac{x2+x1}{2}$ and $\Delta y=y_2-y_1$. The expression of the length of the tether $\langle s_{total}\rangle$ is re-written by replacing equation \ref{eq:sys2_eq4} and \ref{eq:sys2_eq5} into expression \ref{eq:sys2_eq3}, and using the hyperbolic sine properties.
\begin{equation}
\label{eq:cat5}
    s_{total}=2a.sinh(\frac{\Delta x}{a}).cosh(\frac{x_{average}-x_0}{a}),
\end{equation}
Equation \ref{eq:cat6} presents a useful relation between $\langle x_0\rangle$ and $\langle a\rangle$, which results from the division of $\Delta Y$ by $\langle s_{total}\rangle$.
\begin{equation}
\label{eq:cat6}
    x_0=x_{average}-a.tanh^{-1}(\frac{\Delta Y}{s_{total}})
\end{equation}
The insertion of equation \ref{eq:cat6} in equation \ref{eq:cat4} allows to obtain equation \ref{eq:cat7}, and using the Newton-Raphson method on this last produces the value of parameter $\langle a\rangle$. However, when $\Delta Y=0$ it is not possible to compute parameter $\langle a\rangle$ since equation \ref{eq:cat7} does not depend on $\langle a\rangle$ to be valid. Nevertheless, from the knowledge that $\Delta Y=0$ it comes that $x_0$ is known and corresponds to $x_{average}$, which means that equation \ref{eq:cat5} can compute the parameter $\langle a\rangle$. An alternative approach is to force the value of $\Delta  Y$ to be not null by adding a small offset.
\begin{equation}
\label{eq:cat7}
    \Delta Y - 2.a.sinh(\frac{\Delta x}{a}).sinh(tanh^{-1}(\frac{\Delta Y}{s_{total}}))=0
\end{equation}
The expansion of the Taylor series derives the initial estimation of parameter $\langle a\rangle$. 
\begin{equation}
\label{eq:tayl3}
    sinh(x)=\frac{e^x-e^-x}{2}=\sum_{n=0}^{\inf}\frac{x^{2n+1}}{(2n+1)!}=x+\frac{x^3}{3!}+\frac{x^5}{5!}+...
\end{equation}
By applying this approximation to expression \ref{eq:cat7}, one can re-write this last equation as a result of a $ 5^{th}$ order approximation for
the hyperbolic sine, according to equation \ref{eq:tayl4}.
\begin{equation}
\label{eq:tayl4}
    \bigg(\frac{\Delta Y}{2.sinh\big(tanh^-1(\frac{\Delta Y}{s_{tot}})\big)}-\Delta x\bigg)a^4-\frac{\Delta x^3}{3!}a^2-\frac{\Delta x^5}{5!}=0
\end{equation}
The assumption that $\alpha=a^2$ reduces equation \ref{eq:tayl4} to the $2^{nd}$ order. Furthermore, the quadratic formula presented in equation \ref{eq:cat9} produces the solution for this $2^{nd}$ order expression
\begin{equation}
\label{eq:cat9}
    x^2=\frac{-b+/-\sqrt{b^2-4.a.c}}{2.a},
\end{equation}
where 
\begin{equation}
\label{eq:cat10}
    a=\frac{\Delta Y}{2.sinh(tanh^{-1}(\frac{\Delta Y}{s_{total}}))}-\Delta x,
    b= \frac{-\Delta X^3}{3!},
    c= \frac{-\Delta x}{5!}.
\end{equation}
Reverting the variable substitution produces the desired value for $\langle a \rangle$, according to equation \ref{eq:tayl4}. 
\begin{equation}
\label{eq:cat8}
    a=\sqrt(\alpha),
\end{equation}
Since equation \ref{eq:tayl4} is a $4^{th}$ order equation it produces 4 roots -  in the relevant domain, two of them are complex roots and two of them real roots, one positive and other negative. Only the positive real root has a physical meaning and so it is the only one to be taken into account.
\par
The replacement of $\langle x_0\rangle$ and $\langle a\rangle$ into equation \ref{eq:sys2_eq1} or into equation \ref{eq:sys2_eq2} allows to compute the $\langle C\rangle$ parameter.
%%%%%%%%%%%%%%%%%%%%
\subsubsection{Validation of the catenary model}
\begin{figure}[H]
    \centering
    \includegraphics[width=0.9\linewidth]{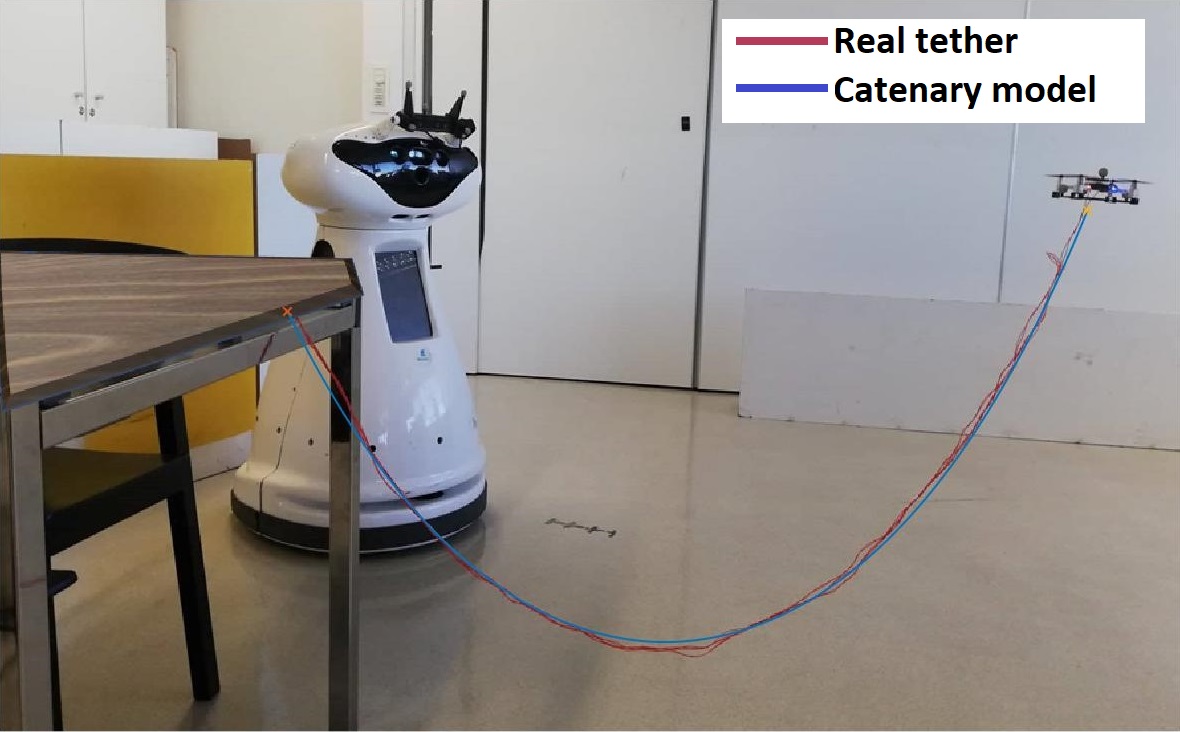}
    \caption{Catenary model validation}
    \label{fig:c_val}
\end{figure}

The malleability of the tether is one of the most important feature to ensure that the tether outlines a catenary curve. The validation of the catenary model was done by overlaying the real silicon tether with the theoretical model of the catenary curve, as shown in figure \ref{fig:c_val}.
%%%%%%%%%%%%%%%%%%%%%%%%%%%%%%%%%%%%%%%%%%%%%%%%%%%%%%%%%%%%%%%%%%%%%%%%%%%%%%
\subsection{Tension Estimation}
\label{sec:tens_estimate}
The tension estimate applied to the quadcopter is computed using the quadcopter's thrust and the inertial information from its IMU sensors. However, these sensor readings present a high level of noise, which means that it is impossible to determine the tension applied to the quadcopter with the desired accuracy. To filter the undesired noise, and assuming that the noise is white Gaussian, a Kalman filter was implemented. 
\par
Equations \ref{eq:sysk1} and \ref{eq:sysk2} describe a linear system, in which $w_k$ and $v_k$ are the process and the observation noise, respectively; $x_k$ is the system's state vector; $y_k$ is the observation vector of the system's states; $u_k$ is the system's input.
\begin{equation}
\label{eq:sysk1}
    x_k=Ax_{k-1}+Bu_k+w_k
\end{equation}
\begin{equation}
\label{eq:sysk2}
    y_k=Cx_k+D+v_k
\end{equation}
The Kalman filter starts by propagating the process model, where $\hat{x}^-_k$ is the state estimate. 
\begin{equation}
     \hat{x}^-_k=A\hat{x}_{k-1}+Bu_k
\end{equation}
Afterwards, it uses the information from the sensor measurements to improve the estimate obtained from the model propagation. The final state estimate is given by:
\begin{equation}
     \hat{x}_k=\hat{x}^-_k+K_k(y_k-C\hat{x}^-_k)
\end{equation}
The variable $K_k$ is the Kalman gain, which adjusts the relation between the process and observation estimates.
\par
The tension that is applied to the quadcopter may be applied through human interaction, which implies infinite possibilities for the way that the wire is pulled. The implemented solution considers a model where the tension remains the same, which can be extended to situations where the wire is not abruptly pulled and do not have considerable oscillations. 
\par
\begin{figure}[H]
    \centering
    \includegraphics[width=0.8\linewidth]{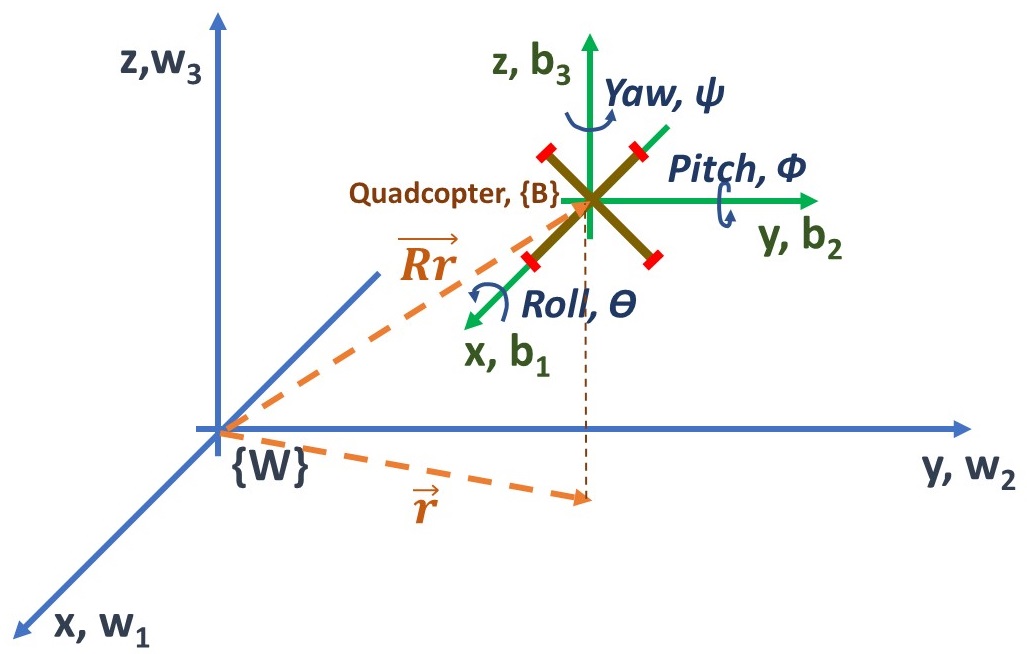}
    \caption{World frame ${W}$, body frame ${B}$, radial vector $\Vec{R}_r$, and its horizontal projection $\Vec{r}$}
    \label{fig:frames}
\end{figure}

The state estimate ($\hat{x_k}$) is a $\Re^3$ vector, including the tension along $x$, $y$ and $z$ directions, and $y_k$ is the observation vector, which includes the tension measurements.
\begin{equation}
\label{eq:Kalman_vars}
\begin{matrix}
    \begin{matrix}
        \hat{x}_k=
    \end{matrix}
    \begin{bmatrix}
        Tx_k \\
        Ty_k \\
        Tz_k \\
    \end{bmatrix}
    y_k=
    \begin{bmatrix}
        Tx_k^{obs} \\
        Ty_k^{obs} \\
        Tz_k^{obs} \\
    \end{bmatrix}
    \\
    \\
    A=C=\begin{bmatrix}
        I
    \end{bmatrix}^{3x3}
    B=D=\begin{bmatrix}
        \varnothing
    \end{bmatrix}^{3x1}
\end{matrix}
\end{equation}
The tension measurements $T^{obs}$ are obtained through equation \ref{eq:NE_tether}, where $\Vec{a}$ corresponds to the acceleration vector, $\Vec{g}$ to the gravity vector, $R(\eta)$ to the rotation matrix between the world and the quadcopter frames (see figure \ref{fig:frames}), and $F_p$ to the total thrust of the quadcopter. The superscript $^{obs}$ refers to the on-board sensor reading of the quadcopter. 
\begin{equation}
\label{eq:NE_tether}
        T^{obs}=m(\Vec{a}^{obs}+\Vec{g})-R(\eta)^{obs}F_p^{obs} + F_{ext}
\end{equation}
%%%%%%%%%%%%%%%%%%%%%%%%%%%%%%%%%%%%%%%%%%%%%%%%%%%%%%%%%%%%%%%%%%%%%%%%%%%%%%
\subsection{Position Estimation}
\label{sec:pos_est}
%%%%%%%%%%%%%%%%%
\label{sec:pos_knowing2points&length}
This section\footnote{In section \ref{sec:t_shape} the catenary model was evaluated for a $\Re^2$ space. In this section, the quadcopter is moving in a $\Re^3$, and so the $y$ and $x$ coordinates in section \ref{sec:t_shape} are replaced by $z$ and $r$ coordinates, where $r=\sqrt{x^2+y^2}$ (see figure \ref{fig:frames})} uses the relation between the tension that the catenary model exerts to the quadcopter and its parameters to compute the position estimate of the quadcopter. The curve parameters $<a>$ and $<s_2>$ are computed according to equations \ref{eq:a_param} and \ref{eq:s2_param}, where the horizontal $H$ and vertical $T_v$ tensions are obtained using the methods presented in section \ref{sec:tens_estimate}.  
\begin{equation}
\label{eq:a_param}
    a=\frac{H}{\omega}
\end{equation}
\begin{equation}
\label{eq:s2_param}
    s_2=\frac{T_v}{\omega}
\end{equation}
The curve parameters $<a>$ and $<s_2>$ are then used to compute the spatial coordinates of the end of the tether attached to the UAV, along with the knowledge of the tether's full length $<s_{tot}>$.
\begin{equation}
\label{eq:posfull_stot}
    s_{tot}=s_2+s_1
\end{equation}
This way, replacing $x_0$ by $r_0$ and $x_1$ by $r_i$ in equation \ref{eq:sys2_eq4}, and using the relation presented in equation \ref{eq:posfull_stot}, one can derive equation \ref{eq:posfull_r_0}.
\begin{equation}
\label{eq:posfull_r_0}
    r_0=r_i+a.sinh^{-1}(\frac{s_{tot}-s_2}{a})
\end{equation}
Furthermore, replacing $x_0$ by $r_0$ and $x_2$ by $r$ in equation \ref{eq:sys2_eq5}, the radial distance $r$ comes as:
\begin{equation}
\label{eq:posfull_radial}
    r=r_0+a.sinh^{-1}(\frac{s_2}{a})
\end{equation}
At last, equation \ref{eq:posfull_z} uses the catenary's expression and computes the quadcopter's altitude,
\begin{equation}
\label{eq:posfull_z}
    z=a.cosh(\frac{r-r_0}{a})+C
\end{equation}
where equation \ref{eq:posfull_C} computes parameter $\langle C\rangle$.
\begin{equation}
\label{eq:posfull_C}
    C=z_i-a.cosh(\frac{r_i-r_0}{a})
\end{equation}
If the horizontal tension $H$ is null, equations \ref{eq:posfull_radial} and \ref{eq:posfull_z} have a mathematical indetermination of type $0\times\infty$. The limits of those equations are presented in equations \ref{eq:posfull_limz} and \ref{eq:posfull_limr}.
\begin{equation}
\label{eq:posfull_limz}
     \lim_{a\to0}z=\lim_{a\to0}a.cosh(\frac{r}{a})+C=z_i+|s_2|-|s_1|
\end{equation}
\begin{equation}
\label{eq:posfull_limr}
     \lim_{a\to0}r=\lim_{a\to0}a.sinh^{-1}(\frac{s_1}{a})+a.sinh^{-1}(\frac{s_2}{a})=0
\end{equation}

%% file: DraftE.tex
\section{Experimental results}
\subsection{Estimation of the vertical tension}
\begin{figure}[h]
    \centering     
    \subfigure[Hovering at a height of 0,3m.]{
        \includegraphics[width=\linewidth]{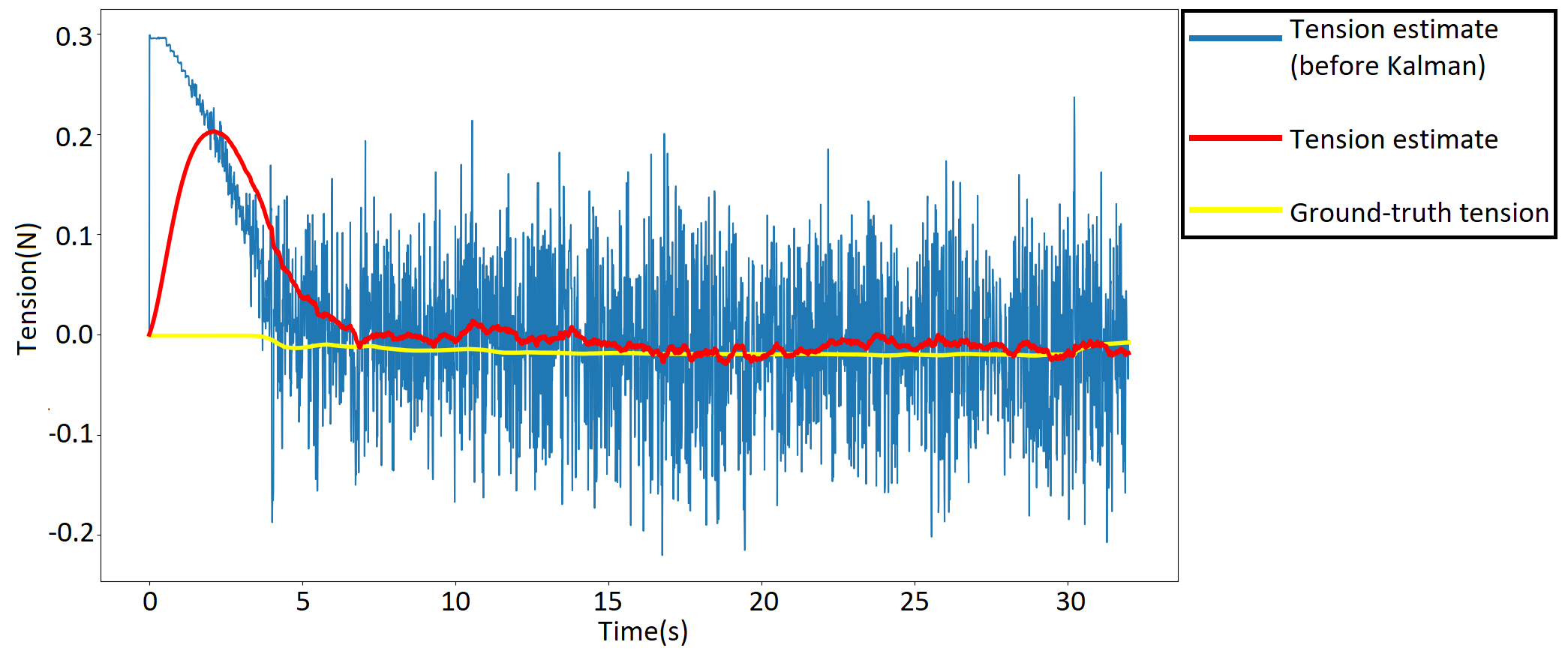}
    }
    \subfigure[Hovering at a height of 1,3m.]{
        \includegraphics[width=\linewidth]{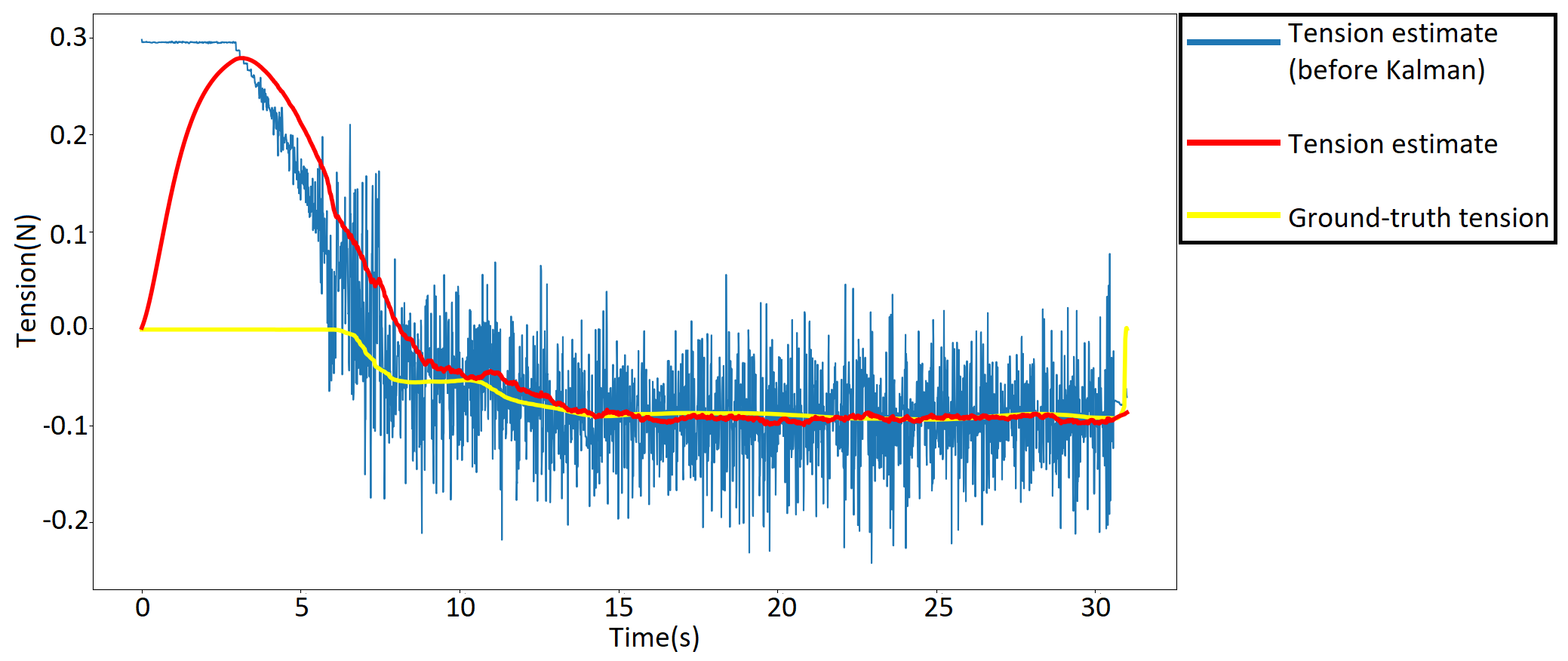}
    }
    \caption{Ground-truth values and estimates of the vertical tension applied to the UAV.}
    \label{fig:exp_vertTension}
\end{figure}

The validation of the vertical tension estimate is done by performing a vertical tethered takeoff, which represents the simplest experiment to compute its ground-truth value $Tv_{gt}$. 
\begin{figure}[h]
    \centering
    \includegraphics[width=0.7\linewidth]{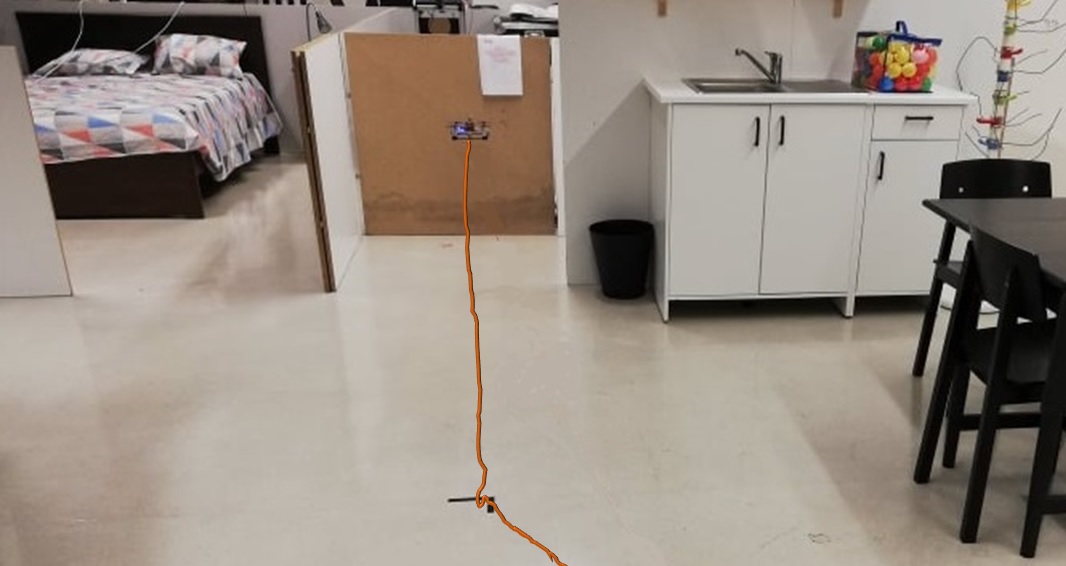}
    \caption{Illustration of the vertical tension estimation experiment}
    \label{fig:my_label}
\end{figure}
This last one is calculated through the height of the quadcopter $z$, and the weight per length unit of the the tether $\omega$.
\begin{equation}
    Tv_{gt}=\omega.z
\end{equation}
Figure \ref{fig:exp_vertTension} illustrates the vertical tension estimates.
The tension estimation procedure does not take into account the force that the ground exerts on the quadcopter while the propellers are not spinning ,i.e., that are not counter balancing its weight. Thus, in the initial instants, the vertical tension estimate does not correspond to its ground-truth values and the 0,3N value approximately corresponds to the weight of the quadcopter.
%%%%%%%%%%%%%%%%%%%%%%%%%
\subsection{Estimation of the horizontal tension}
\begin{figure}[htb]
    \centering
    \includegraphics[width=0.8\linewidth]{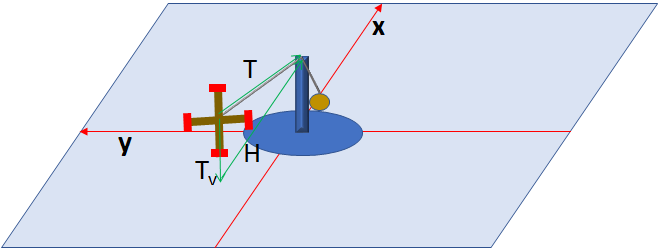}
    \caption{Test-bench for computing the horizontal tension.}
    \label{fig:Hor_test_bench}
\end{figure}

\begin{figure}[htb]
    \centering
    \subfigure[Hovering flight at position $x=-0.3$, $y=z=0$.]{
        \includegraphics[width=0.8\linewidth]{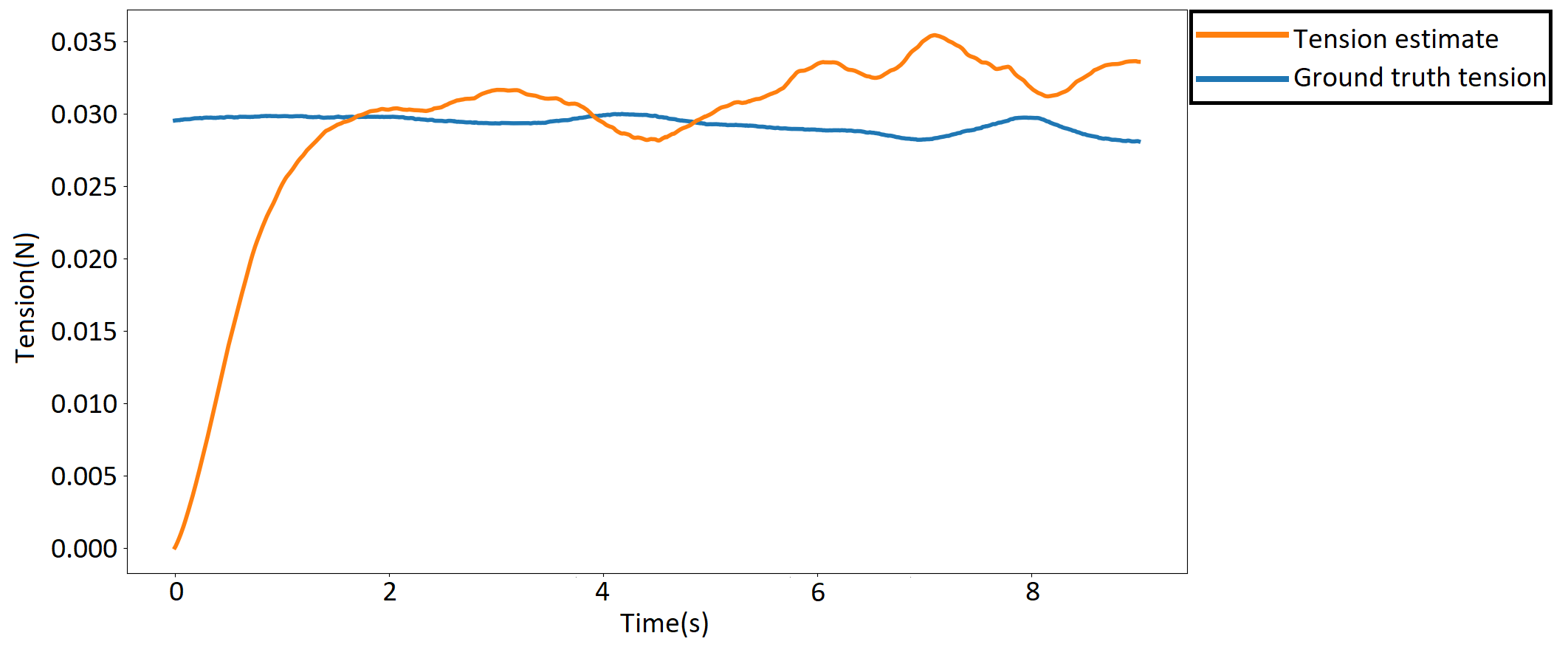}
        \label{subfig:T_hor1}
    }
    \subfigure[Hovering flight at $x=y=0$ and $z=0.4$.]{
        \includegraphics[width=0.8\linewidth]{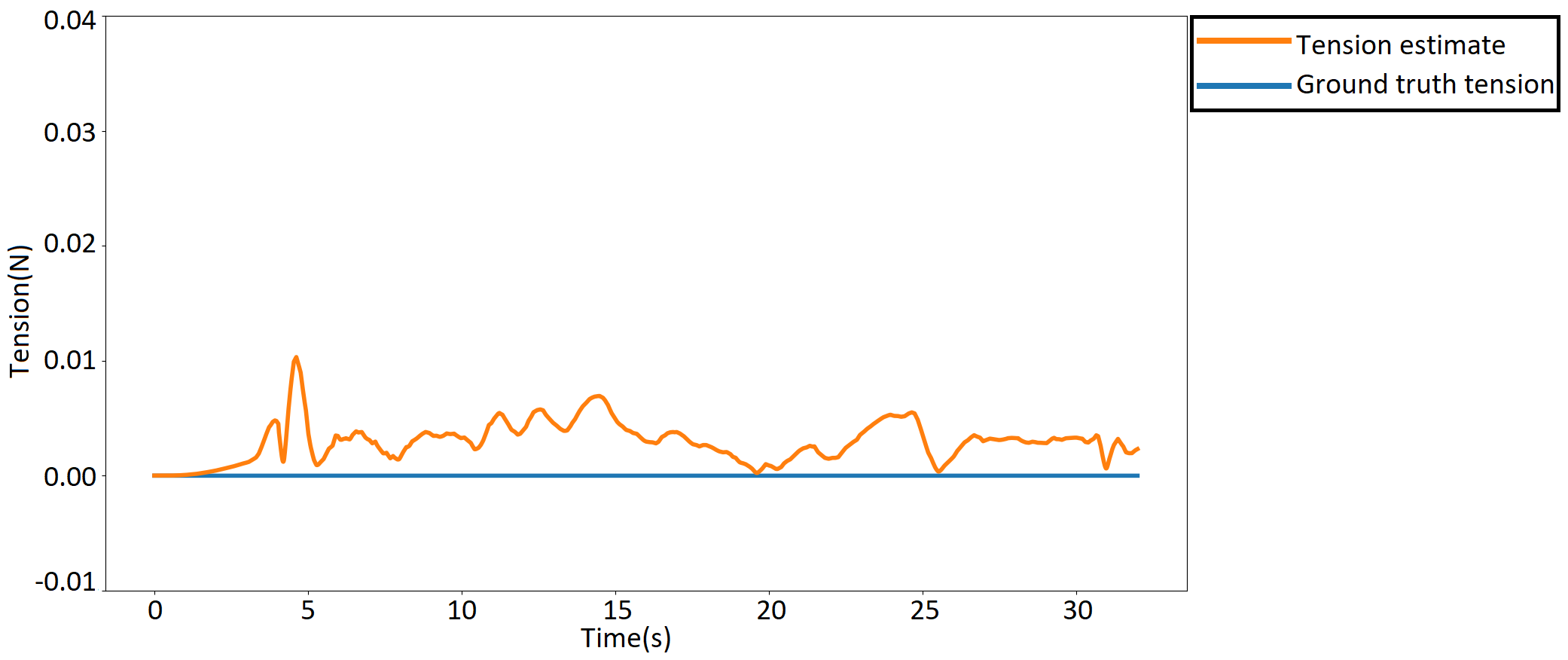}
        \label{subfig:T_hor2}
    }
    \caption{Ground-truth values and estimates of the horizontal tension applied to the UAV.}
    \label{fig:Hor_sfio}
\end{figure}

To validate the horizontal tension estimate, its ground-truth value is computed using a small mass (coin) attached to a wire. Figure \ref{fig:Hor_test_bench} illustrates the scheme of the test-bench used to compute the ground-truth values for the horizontal tension.
Equation \ref{eq:exp_HorGroundThruth} computes the ground-truth value of the horizontal tension, in which T corresponds to the weight of the mass and $\gamma$ is computed according to equation \ref{eq:exp_gamma} - $z_q$ and $z_a$ are the height of the quadcopter and the height of the vertical arm, respectively, and $r_q$ is the quadcopter's radial coordinate, assuming the inertial frame presented in figure \ref{fig:Hor_test_bench}.
\begin{equation}
    \label{eq:exp_HorGroundThruth}
    H=cos(\gamma).T
\end{equation}
\begin{equation}
\label{eq:exp_gamma}
    \gamma=tan^-1(\frac{z_q-z_a}{r_q})
\end{equation}
Figure \ref{subfig:T_hor1} displays the tension estimate and the ground-truth value for an attached mass of 4,1g, and figure \ref{subfig:T_hor2} shows the result for a tethered vertical takeoff, where the tether direction is mainly vertical.
As figure \ref{subfig:T_hor2} displays, the horizontal tension estimate is nearly null - less than 0.01N.
%%%%%%%%%%%%%%%%%%%%%%%%%%%%%%%%%%%%%%%%%%%%%%%%%%%%%%%%%%%%%%%%%%%%%%%%%%%%%%%%
\subsection{Tension following}
The tension applied to the quadcopter is estimated in real-time, according to section \ref{sec:tens_estimate}. The goal position of the quadcopter changes if the estimated tension is greater than a pre-defined threshold. When this occurs, the goal position of the quadcopter is successively updated to its current position, making it to follow the pull's direction. When the tension applied to the quadcopter is no longer greater than the pre-defined threshold, the goal position stops being updated and the quadcopter remains hovering at its last position.
\par
\begin{figure}[htb]
    \centering
    \includegraphics[width=\linewidth]{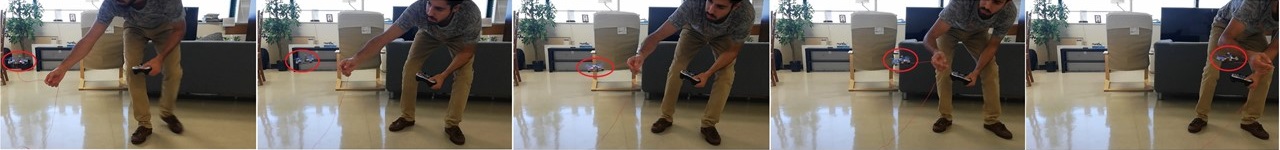}
    \caption{Screenshots of the quadcopter following the tension's direction.}
    \label{fig:tens_foll_vid1}
\end{figure}
\begin{figure}[H]
    \centering
    \includegraphics[width=\linewidth]{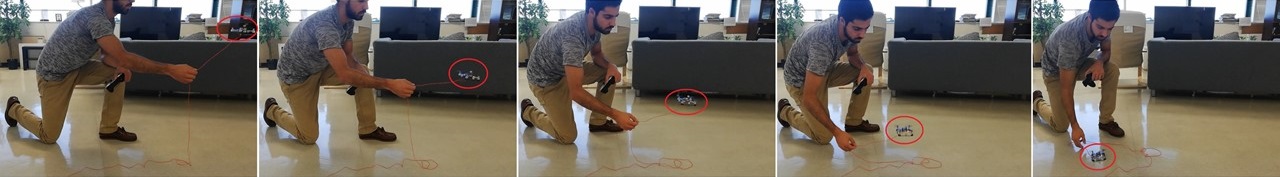}
    \caption{Screenshots of the landing of the quadcopter using the \textit{tension following} feature.}
    \label{fig:tens_foll_vid2}
\end{figure}

To observe the behaviour of the \textit{tension following} feature a few videos were taken \footnote{The full videos are presented on Youtube \href{https://www.youtube.com/playlist?list=PLqlQHq20tO7XfLuxZbgtqlhNLZgoq9_cl}{\underline{here}}.}, where figures \ref{fig:tens_foll_vid1} and \ref{fig:tens_foll_vid2} correspond to screenshots of those videos.
Moreover, the implementation of the \textit{tension following} feature can also be used for the landing process. Given that, after the first tug, a flag is activated indicating that the \textit{tension following} feature is on. Thus, if the quadcopter flies under a certain height, the motors are turned off. Without using an external motion system, the same principle can be applied using a sensor distance, which deactivates the motors if the distance to the landing platform is smaller than a threshold. Figure \ref{fig:tens_foll_vid2} illustrates the mentioned landing process.
%%%%%%%%%%%%%%%%%%%%%%%%%%%%%%%%%%%%%%%%%%%%%%%%%%%%%%%%%%%%%%%%%%%%%%%%%%%%%%%%
\subsection{Position estimation}
\begin{figure}[h]
    \centering
    \subfigure[Hovering at $x\simeq1,2m$, $y\simeq0m$ and $z\simeq1m$.]{
        \includegraphics[width=0.9\linewidth]{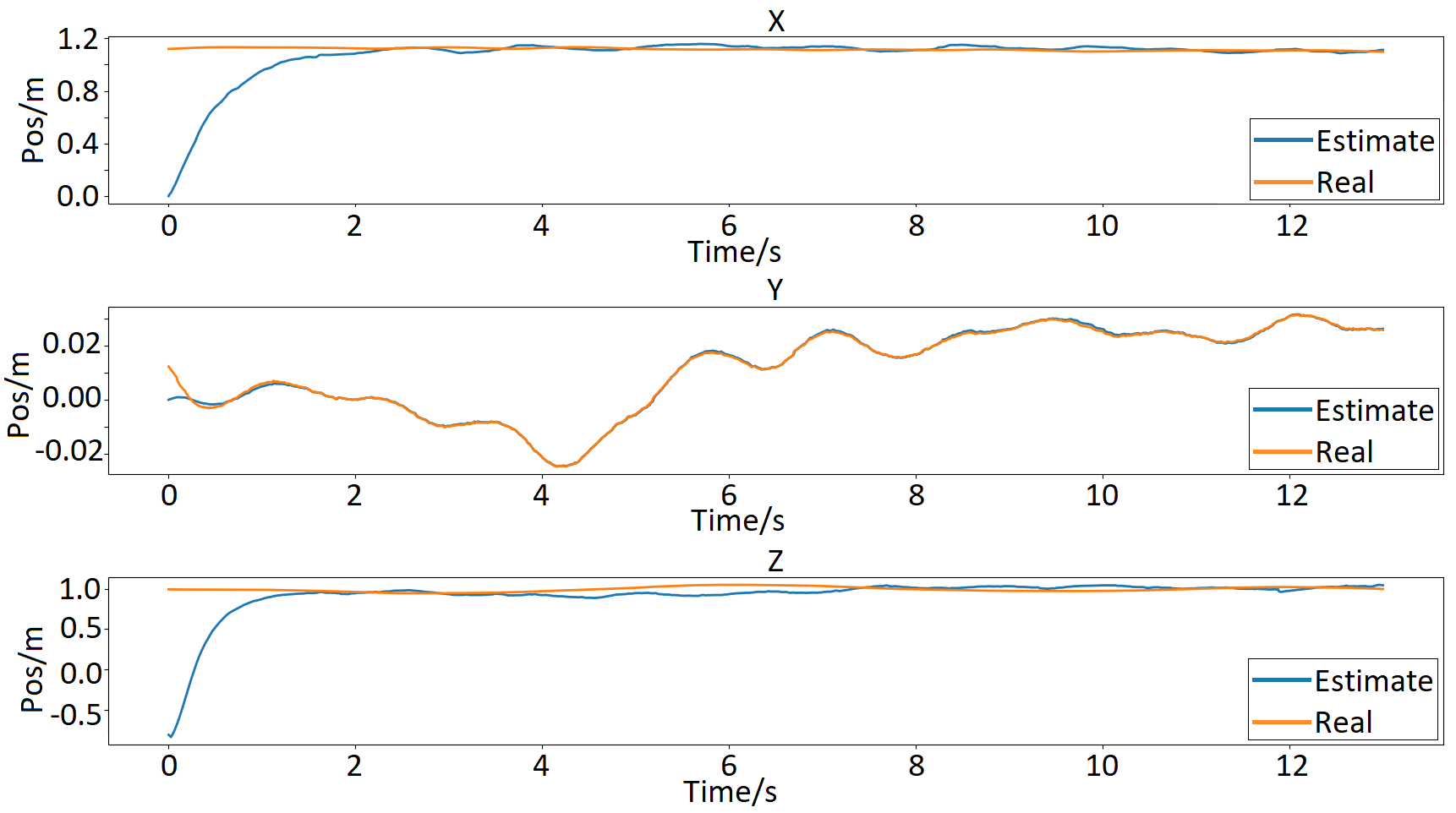}
        \label{subfig:y_high1}
    }
    \subfigure[Hovering at $x\simeq1m$, $y\simeq-0.3m$ and $z\simeq1m$.]{
        \includegraphics[width=0.9\linewidth]{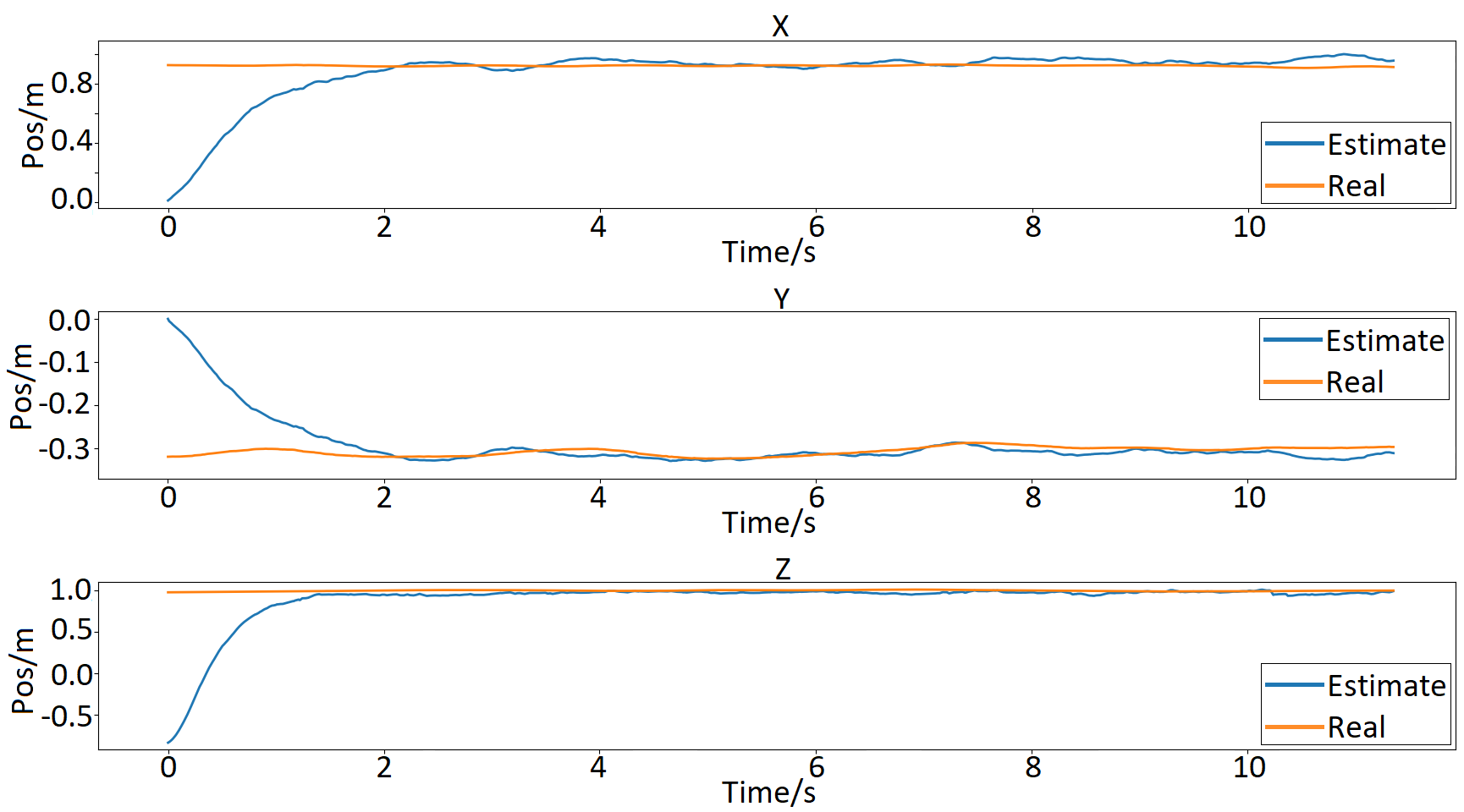}
    }
    \caption{Hovering at same height $z$ and similar radial distance ($x$, $y$) for small angle $\beta$ ($<20º$).}
    \label{fig:expPos_pair1}
\end{figure}

\begin{figure}[h]
    \centering
    \subfigure[Hovering at $x\simeq1m$, $y\simeq0m$ and $z\simeq1,2m$.]{
        \includegraphics[width=0.9\linewidth]{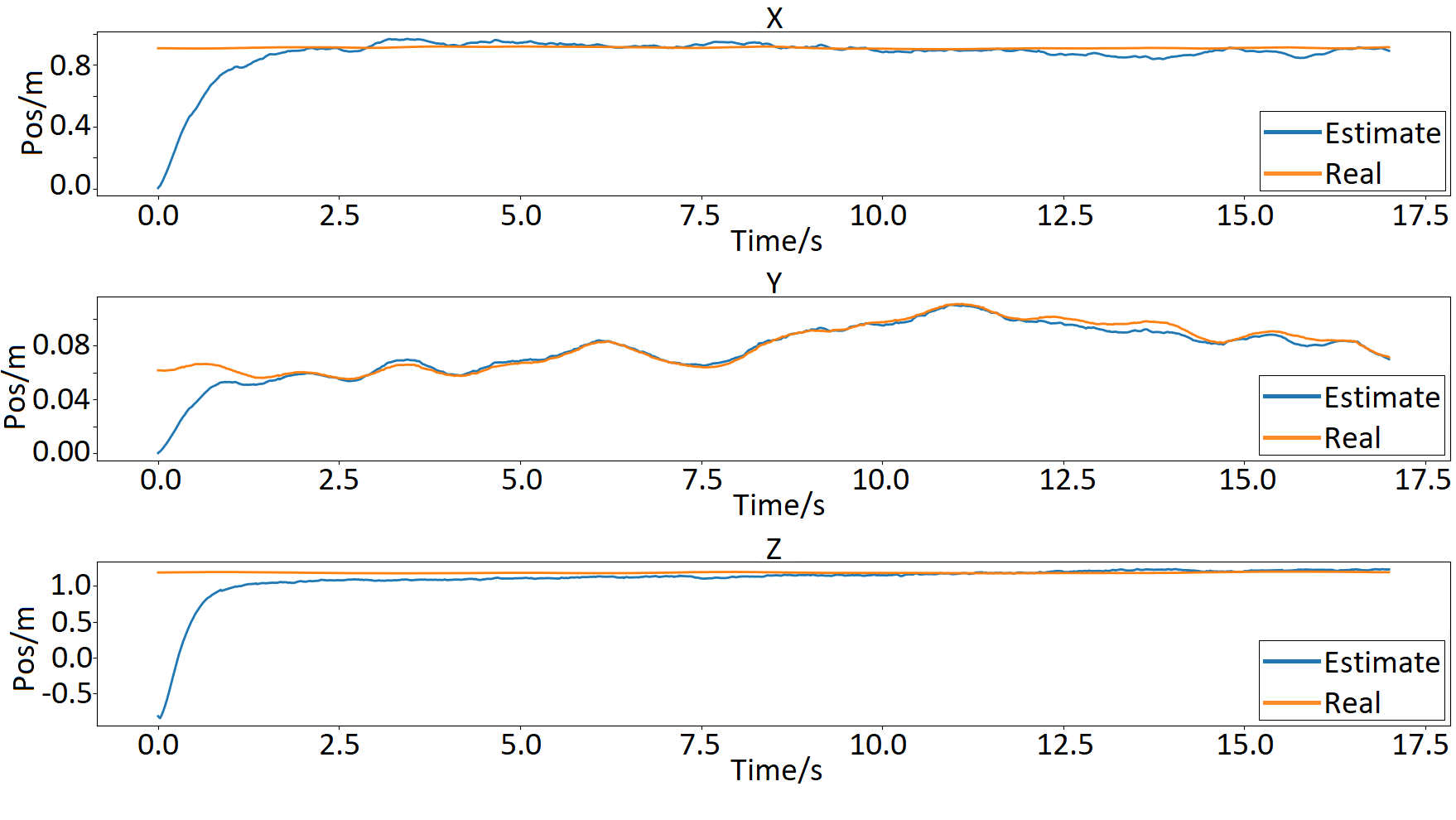}
        \label{fig:pos_est_z12}
    }
    \subfigure[Hovering at $x\simeq1m$, $y\simeq-0.3m$ and $z\simeq0,5m$.]{
        \includegraphics[width=0.9\linewidth]{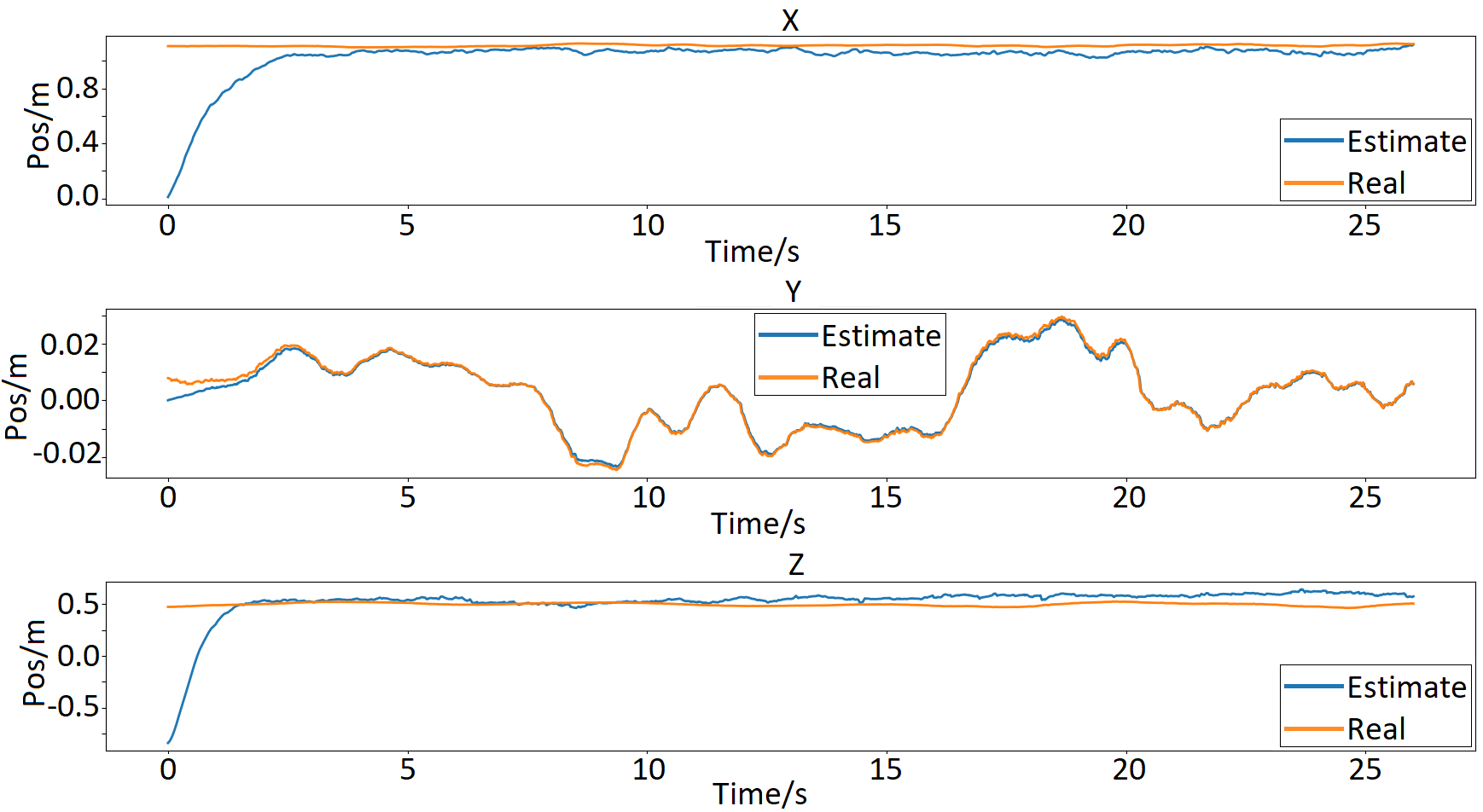}
        \label{subfig:y_high2}
    }
    \caption{Hovering at different heights $z$ and similar radial distance ($x$, $y$) for small angle $\beta$ ($<10º$).}
     \label{fig:expPos_pair2}
\end{figure}

\begin{figure}[h]
    \centering
    \subfigure[Hovering at $x\simeq0,5m$, $y\simeq-0.6m$ and $z\simeq1m$.]{
        \includegraphics[width=0.9\linewidth]{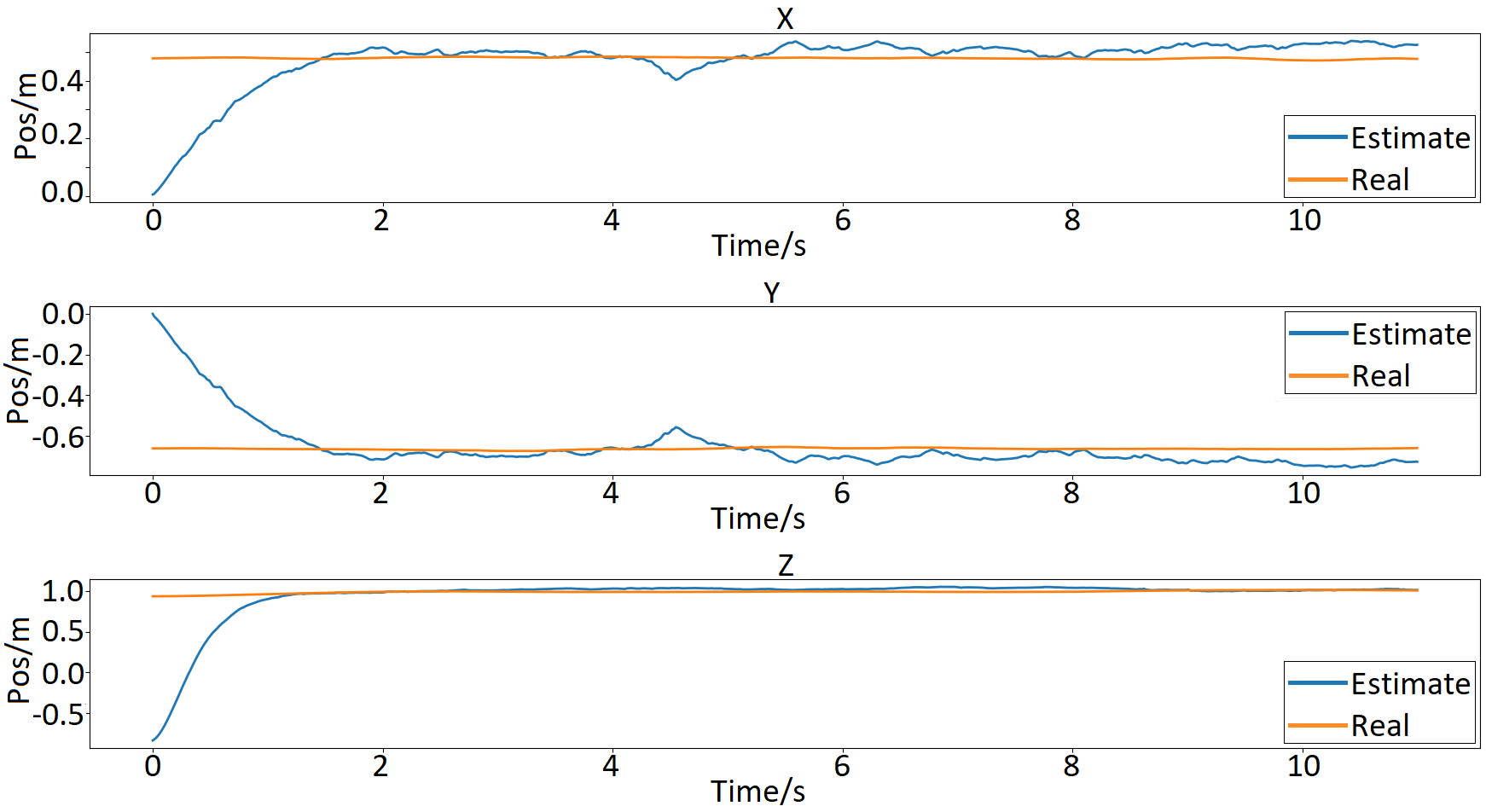}
        \label{fig:pos_est_yneg1}
    }
    \subfigure[Hovering at $x\simeq0,9m$, $y\simeq0.6m$ and $z\simeq1m$.]{
        \includegraphics[width=0.9\linewidth]{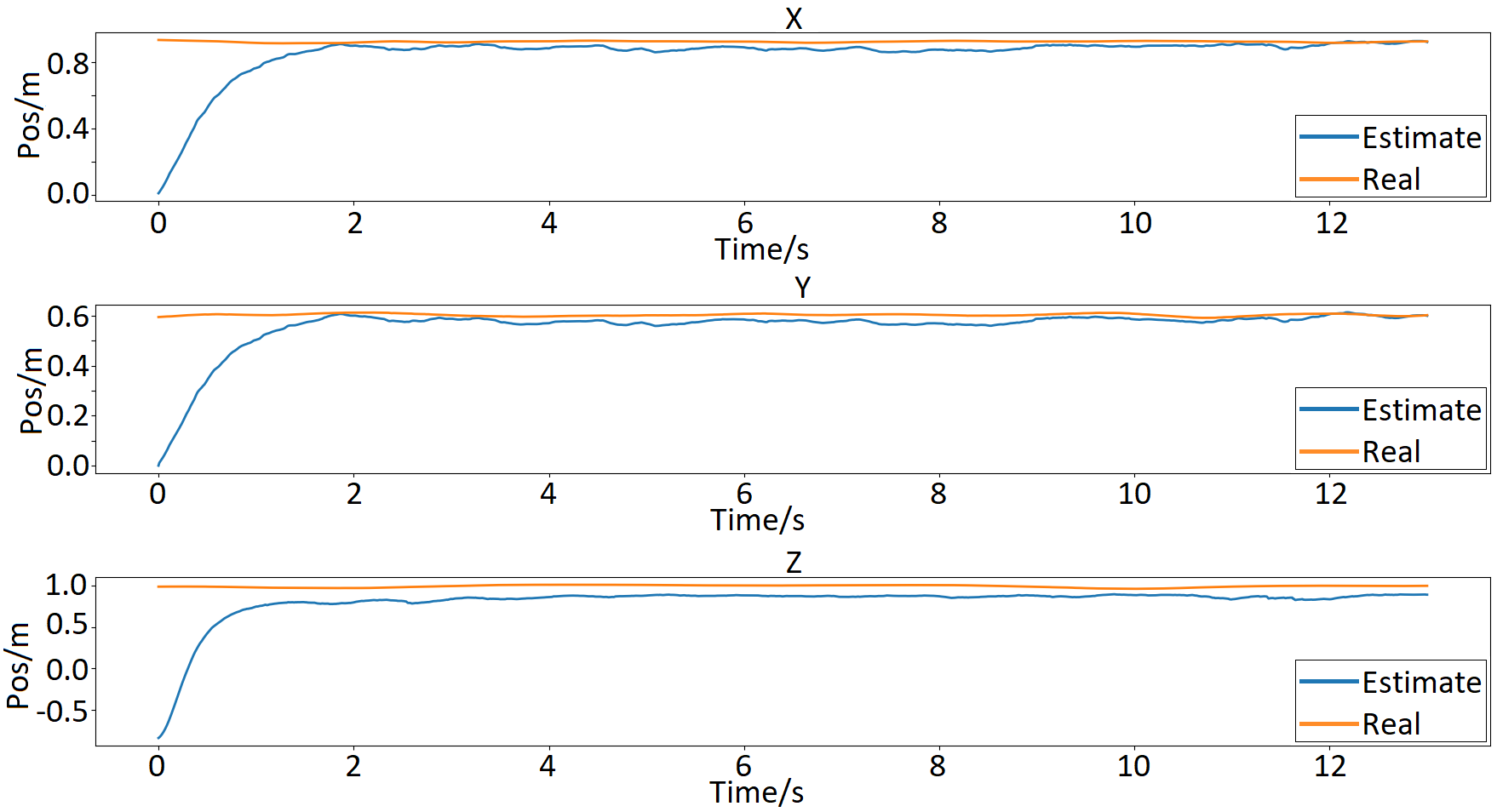}
    }
     \caption{Hovering at same height $z$ and different radial distance ($x$, $y$) for non-small angle $\beta$ ($>30º$).}
    \label{fig:expPos_pair3}
\end{figure}

To avoid the modeling of the tether's oscillations, the results presented throughout this section concern hovering flights. Additionally, the angle $\beta$, which relates the radial distance with the $x$ and $y$ coordinates, was also a source of inaccuracy. It was initially assumed that the $\beta$ angle was known and was computed using the Mocap system. In practice, the angle $\beta$ could be computed using a visual or mechanical system on the ground controller that could indicate the direction of the tether. Figure \ref{fig:expPos_pair1} presents a pair of experiments in which the height corresponds to 1m and the radial distances are similar between them.
In a second set of experiments (figure \ref{fig:expPos_pair2}) the radial distances are also similar between them but the estimation of the quadcopter's height is evaluated for two different heights - 0,5m and 1,2m.
A third set of experiments (figure \ref{fig:expPos_pair3}) is performed with a bigger range for the value of the $y$ coordinate.
In the initial instants the Kalman filter assumes that the tensions $T_x$ and $T_y$ are null, which means that the length $\langle s_2\rangle$ and the tether parameter $\langle a\rangle$ also start to be zero (see equations \ref{eq:expPos_a} and \ref{eq:expPos_s2}).
\begin{equation}
\label{eq:expPos_a}
    s_2=\frac{T_v}{\omega}
\end{equation}
\begin{equation}
\label{eq:expPos_s2}
    a=\frac{H}{\omega}
\end{equation}
According to the expression deduced in equation \ref{eq:posfull_limr}, the initial estimated position of the radial coordinate is null, implying that $x$ and $y$ coordinates are also null. On the one hand, equation \ref{eq:posfull_limz} allows to infer that the initial estimate regarding the altitude corresponds to $z_i-|s_1|$, since $s_2$ is zero. On the other hand, the tether's total length is given by equation \ref{eq:posfull_stot}, which means that $s_1=s_{tot}$ for a null length $s_2$. Equation \ref{eq:in_instants} presents the initial estimate of the altitude over this set of experiments. 
\begin{equation}
\label{eq:in_instants}
    z=z_i-|s_1|=0,754-1,6=-0,846
\end{equation}

%% file: 8conclusion.tex
\section{Conclusions}
This work presented a method to estimate the tension applied to a quadrotor by using measurements from the IMU sensors. Due to the high level of noise, two Kalman based filtering processes were introduced. Furthermore, the tension estimate was used to present alternative ways of controlling the quadcopter, and to improve the position estimate of a tethered quadcopter.
\par
The first flight control strategy used the tension estimate to update the quadcopter's position. Nevertheless, the position of the quadcopter must be known through an external motion system. Aiming to develop a control strategy that does not need to know the position of the UAV, a novel methodology that uses the tether's shape and the tension estimate was introduced.
\par
Moreover, this study presented a method to estimate the position of the quadcopter based on the tension that the tether applies to the UAV and its shape.